\documentclass{article}                                                

\usepackage{verbatim}
\usepackage{amsmath}
\usepackage{graphicx}
\usepackage{amssymb}
\usepackage{array}
\usepackage[draft]{fixme}
\usepackage[english]{babel}
\usepackage{subfig}

\usepackage{color}

\def\R{\mathbb{R}}

\newcommand{\frm}[1]{\langle #1\rangle}

\newtheorem{theorem}{Theorem}

\def\dist{l}
\def\ty{\tilde\dist}
\def\te{\tilde \theta}
\def\cte{\cos(\te)}
\def\ste{\sin(\te)}
\def\cs{c(s)}

\def\wt{\tilde \omega}

\title{Hybrid Feedback Path Following for Robotic Walkers via
  Bang-Bang Control Actions}

\author{Stefano Divan and Daniele Fontanelli and Luigi
  Palopoli 
  \thanks{This project has received funding from the European Union’s
    Horizon 2020 Research and Innovation Programme - Societal
    Challenge 1 (DG CONNECT/H) under grant agreement n$^\circ$ 643644
    ``ACANTO - A CyberphysicAl social NeTwOrk using robot
    friends''.}
  \thanks{D. Fontanelli is with the Department of Industrial
    Engineering (DII), University of Trento, Via Sommarive 5, Trento,
    Italy {\tt\small daniele.fontanelli@unitn.it}. S. Divan and
    L. Palopoli are with the Department of Information Engineering and
    Computer Science (DISI), University of Trento, Via Sommarive 5,
    Trento, Italy {\tt\small \{palopoli,divan\}@disi.unitn.it}}%
}

\begin{document}

\maketitle
\thispagestyle{empty}
\pagestyle{empty}

\begin{abstract}
  We show a control algorithm to guide a robotic walking assistant
  along a planned path. The control strategy exploits the
  electromechanical brakes mounted on the back wheels of the walker.
  In order to reduce the hardware requirements we adopt a 
  Bang Bang approach relying of four actions (with saturated value for the
  braking torques).
  
  When the platform is far away from the path, we execute an approach
  phase in which the walker converges toward the platform with a
  specified angle.  When it comes in proximity of the platform, the
  control strategy switches to a path tracking mode, which uses the
  four control actions to converge toward the path with an angle which
  is a function of the state.  This way it is possible to control the
  vehicle in feedback, secure a gentle convergence of the user to the
  planned path and her steady progress towards the destination.
\end{abstract}


\section{Introduction}

The use of robotic platforms to help older adults navigate complex
environments is commonly regarded as an effective means to extend
their mobility and, ultimately, to improve their health conditions.
The EU Research project ACANTO~\cite{ACANTO} aims to develop a robotic
friend (called {\em FriWalk}), which offers several types of cognitive
and physical support.  The {\em FriWalk} looks no different from a
classic rollator, a four wheel cart used to improve stability and
receive physical support. However, its sensing and computing abilities
allow the {\em FriWalk} to sense the environment, localise itself and
generate paths that can be followed with safety and comfort. The user
can be guided along the path using a Mechanical Guidance Support
(MGS).

The MGS utilises electromechanical brakes to steer the vehicle in
order to stay as close as possible to the planned path. In this paper,
we study a control strategy that fulfills this goal.  A few specific
issues makes our problem particularly challenging.  First, the low
target cost of the device prevents us from using expensive sensors to
estimate the force and the torque applied by the user to the
platform. Second, the use of electromechanical brakes and the
sometimes difficult grip conditions make the braking action difficult
to modulate.  Third, because the guidance system interacts with the
user (who ultimately pushes the vehicle forward), it should be
flexible to adapt to different users and different operating
conditions and it should give directions that users should find
``sensible'' and easy to follow.

This being said, we propose a strategy that operates in two different
ways when the user is far away from the path and when she is in its
proximity. In this first condition, the controller seeks to ``approach the path'', in the 
second condition it seeks to ``follow the path''.

For the approach phase, the controller reduces the distance from the
path by executing ``a few'' control actions, which are easily
understood by the user who can be left in control of her motion for
most of the time. A useful inspiration to design the controller for
this phase can be found in the work of Ballucchi et
al.~\cite{BicchiBBC96}. These authors proved that for a vehicle that
moves with constant speed and with limited curvature, the control
policy that takes the vehicle to the path in minimum time is a Bang
Bang strategy with saturated controls.  In our case, this means
restricting to four control actions: 1. let the user go, 2. force a
right turn blocking the right wheel, 3. force a left turn blocking the
left wheel, 4. force the complete halt of the system blocking both
wheels. The resulting motion of the vehicle is given by a
concatenation of straight lines and circles with fixed radius.  The
restriction to a Bang Bang strategy based on saturated is convenient
for us because it does not require any force measurements and or
finely modulated braking actions. However, the minimum time manoeuvres
could appear unusual and uncomfortable to the user. For this reason,
our control strategy for the approach phase uses saturated control
\emph{but} allows us to specify the angle of approach (the angle
between the orientation of the vehicle and the tangent to the
trajectory).  This allows us to strike different trade-offs between the
total distance covered and the smoothness of the approach manoeuvres.

When the system comes in proximity of the path, we switch into path
tracking mode.  We still use the same set of control actions, but the
angle of approach is given by a functions of the system's state.  We
show that this produces a stabilising control law.  The key advantage
of the approach is that it allows us to emulate the behaviour of fine
grained control strategies requiring the measurement of the user's
torques, by using a coarse discrete valued actuation and the only
information of the displacement and of the orientation of the platform
with respect to the path.  We prove the efficacy of our solution by
extensive simulations.

The paper is organised as follows. In Section~\ref{sec:relw} we offer
a quick survey of the related work. In Section~\ref{sec:definition} we
introduce the most important definitions and state the problem in
formal terms. In Section~\ref{sec:solution}, we describe our control
strategy and in Section~\ref{sec:results} we show its performance by
means of simulations. In Section~\ref{sec:conclusions} we state our
conclusions and announce future work directions.

\section{Related work}
\label{sec:relw}

Motion control of autonomous vehicles has been widely studied in the
last years. Solutions to path planning and path following have been
provided for a complete spectrum of robotic vehicles, like unicycles
and car-likes, and can be found in many
textbooks~\cite{Laumond98}. The control approaches underpinning such
solutions range in the wide realm of nonlinear control, e.g., chained
forms~\cite{Samson95}, closed loop steering via Lyapunov
techniques~\cite{AicardiBB95} or exponential control
laws~\cite{SordalenW92}.

An important research area concerns the class of path following
problems for nonholonomic vehicles with limited curvature radius or
saturated inputs, which plays a relevant role for assistive robotic
vehicles~\cite{hirata2007motion,FontanelliGPP13,martins2015review}.
This topic has been widely studied in the literature using, for
example, chained forms~\cite{de1998feedback}, Lyapunov-based control
synthesis on the kinematic~\cite{jiang2001saturated} or
dynamic~\cite{LapierreS07,SoetantoLP2003} models or hybrid
automata~\cite{balluchi2005path}.

The device adopted in this paper can be modeled as a particular
Dubins car with a fixed curvature radius. One of the first solution
of driving a Dubins car along a given path has been introduced
by~\cite{BicchiBBC96}. The proposed solution is based on a
discontinuous control scheme on the angular velocity of the
vehicle. This approach has been further developed in~\cite{BicchiBS01}
for optimal route tracking control minimising the approaching path
length.  The optimal problem is based on the definition of a switching
logic that determines the appropriate state of a hybrid system.  From
the same authors, an optimal controller able to track generic paths
that are unknown upfront, provided some constraints on the path
curvature are satisfied~\cite{balluchi2005path}. This second solution
considers the curvature of the path as a disturbance which has to be
rejected.

For what concerns assistive carts, 
the passive walker proposed by
Hirata~\cite{hirata2007motion} is a standard walker, with two caster
wheels and a pair of electromagnetic brakes mounted on fixed rear
wheels, which is essentially the same configurations that we consider
in this paper.  The authors propose a guidance solution using
differential braking, which is inspired to many stability control
systems for cars~\cite{pilutti1995vehicle}.  By suitably modulating
the braking torque applied to each wheel, the walker is steered toward
a desired path, as also reported in~\cite{saida2011development}.

The walking assistant considered in this paper builds atop the model
proposed in~\cite{hirata2007motion, saida2011development}.  In our
previous work~\cite{FontanelliGPP13}, we proposed a control algorithm
based on the solution of an optimisation problem which minimises the
braking torque.  However, the control law relies on real--time
measurements of the torques applied to the walker, which are difficult
without expensive sensors.  Due to cost limitations and simplicity in
the algorithm design, we restrict the possible actions to turn left or
right by blocking alternatively the left or right wheel, thus casting
the problem to the class of path following problems for nonholonomic
vehicles with limited curvature radius.  For example, the {\em
  i-Walker} rollator~\cite{cortes2010assistive} is equipped with
triaxial force sensors on the handles to estimate the user applied
forces.  Nevertheless, its costs is on the order of some thousands
euros, which make it unaffordable for the majority of the potential
users.  In light of this choice, this paper represents a first attempt
to fuse hybrid control laws conceived for optimal tracking of
unicycle-like vehicles~\cite{balluchi2005path} with the nonlinear
trajectory tracking approaches proposed, for instance,
in~\cite{SoetantoLP2003,jiang2001saturated}.  To this end, we first
generalise the hybrid control law proposed in~\cite{balluchi2005path}
to generic and customisable angles of approach to the reference path.
Customisation comes as a degree of freedom that turns out to be very
useful for the problem at hand by determining the angle as a function
of the environmental situation, e.g., to avoid obstacles or leading
the AP as fastest as possible to the reference trajectory.  Using a
time varying angle of approach we are also able to smooth the
convergence to the path as desired.  It has to be noted that with
respect to other existing solutions, e.g.,
\cite{hirata2007motion,FontanelliGPP13}, the proposed solution
simplifies the control algorithm by renouncing to the estimates of the
user applied forces (with remarkable savings in terms of cost).

\section{Background and Problem Formulation}
\label{sec:definition}
 
The device considered in this paper is derived from a commercial
walker inserting electro-mechanical brakes on the rear wheels along
with other mechatronic components.  The {\em FriWalk}
localisation~\cite{NazemzadehFMRP13,NazemzadehMFMP15} is based on
incremental encoders mounted on the back wheels, on an inertial
platform measuring the vehicle accelerations and angular velocities,
and on exteroceptive sensors (RFID readers and cameras).  The vehicle
uses vision technologies to detect information on the surrounding
environment and to plan the safest course of action for the
user~\cite{colombo2013motion,ColomboFLPS15}.  Due to the described
abilities, the reference path is assumed to be known up-front and its
localisation is considered solved.

\subsection{Vehicle Dynamic Model}

\begin{figure}[t]
   \centering
   \includegraphics[width=0.6\columnwidth]{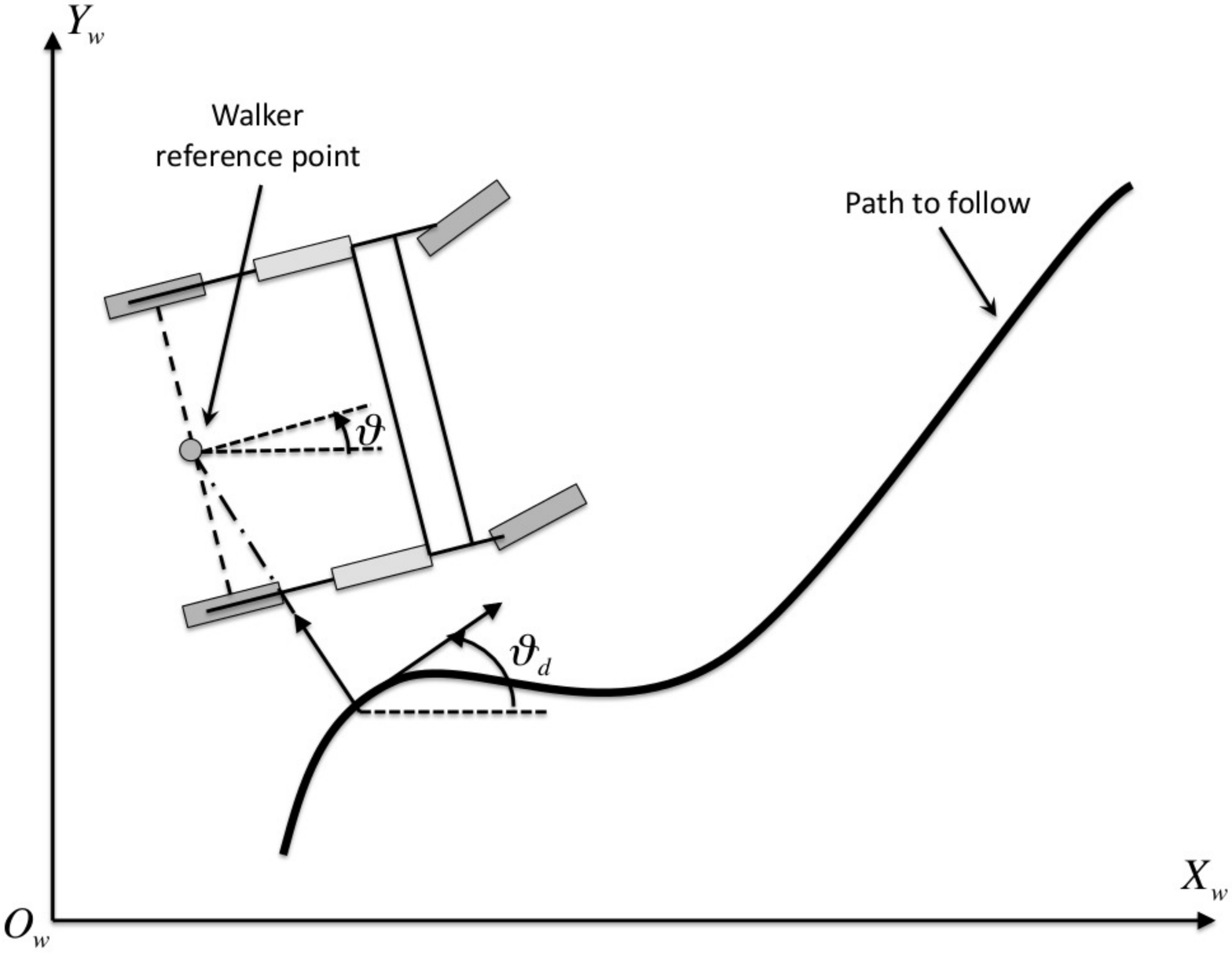}
   \caption{Vehicle to path configuration and reference frames.}
   \label{fig:rf}
\end{figure}

With reference to Fig.~\ref{fig:rf}, let $\frm{W}
=\{O_w,\,X_w,\,Y_w,\,Z_w\}$ be a fixed right-handed reference frame,
whose plane $\Pi = X_w\times Y_w$ is the plane of motion of the cart,
$Z_w$ pointing outwards the plane $\Pi$ and let $O_w$ be the origin of
the reference frame.
Let ${\bf x} = [x,\, y,\, \theta]^T\in \R^2\times S$ be the kinematic
configuration of the cart, where $(x,\, y)$ are the coordinates of the
mid--point of the rear wheels axle in $\Pi$ and $\theta$ is the
orientation of the vehicle w.r.t. the $X_w$ axis (see
Fig.~\ref{fig:rf}). The dynamic model of the {\em FriWalk} can be
assimilated to a unicycle
\begin{equation}
  \label{eq:kin}
  \begin{bmatrix}
   \dot x \\
   \dot y \\
   \dot \theta \\
   \dot v \\
   \dot \omega
 \end{bmatrix} =
  \begin{bmatrix}
   v \cos(\theta) \\
   v \sin(\theta) \\
   \omega \\
   \frac{F}{m} \\
   \frac{N}{J}
 \end{bmatrix} \Leftrightarrow 
  \begin{bmatrix}
    \dot s \\
    \dot \dist \\
    \dot \te \\
    \dot v \\
    \dot \wt
  \end{bmatrix} = 
  \begin{bmatrix}
   \gamma \\
    v \ste \\
    \omega - \cs \gamma \\
    \frac{F}{m} \\
    \frac{N}{J} - \cs \ddot s - \cs' \gamma
  \end{bmatrix}
\end{equation}
where $v$ is the forward velocity of the vehicle and $\omega$ its
angular velocity.  $F$ is the external force acting on the vehicle
along the direction of motion, $N$ is the external torque about the
$Z_w$-axis, $m$ and $J$ are the mass and moment of inertia of the
cart.  Moreover, $s$ is the curvilinear abscissa along the path,
$\dist$ is the distance between the origin of the Frenet frame and the
reference point of the {\em FriWalk} along the $Y$-axis of the Frenet frame,
and $\theta_d$ is the angle between the $X_w$-axis and the $X$-axis of
the Frenet frame (see Fig.~\ref{fig:rf}). Therefore, $\te = \theta -
\theta_d$ and $\wt = \dot \te$.  Furthermore, the path curvature is
defined as $\cs = {d \theta_d(s)}/ {ds}$, while $c'(s) = d \cs /
ds$. Finally,
\[
\gamma = \frac{v \cte}{1-\cs \dist} .
\]
It is worthwhile to note that this model is commonplace in the
literature~\cite{de1998feedback,lapierre2007nonlinear,SoetantoLP2003}.

By denoting with $\tau_r$ and $\tau_l$ the torques applied on the
right and left rear wheels, respectively, we have the invertible
linear relations
\begin{equation}
  \label{eq:f2t}
  F = \frac{\tau_r + \tau_l}{r}  \,\,\,\mbox{and}\,\,\,
  N = \frac{(\tau_r - \tau_l)d}{2r} .
\end{equation}
where $d$ is the axle length.  In particular, denoting with
$M_{(\cdot)}$ the quantity $M$ of the left or right side of the
trolley and assuming that $\alpha_{(\cdot)}$ is the rotation angle of
the rear wheels, the wheel dynamics can be described by $J_w \ddot
\alpha_{(\cdot)} = \tau_{(\cdot)}$, where $J_w$ is the equivalent
moment of inertia of the wheel.

\subsection{Braking System}

The available input variables are the independent braking torques
$\tau_l^b$ and $\tau_r^b$, acting on the left and right wheels,
respectively.  Similarly
to~\cite{hirata2007motion,saida2011development}, the braking torques
ranges between zero, i.e., the wheel rotates freely, to a maximum
value.  More precisely, let the torques acting on~\eqref{eq:f2t} be
expressed by
\begin{equation}
  \label{eq:WheelDynFinal3}
  \tau_{(\cdot)} = \tau_{(\cdot)}^h + \tau_{(\cdot)}^b - b_{w} \dot \alpha_{(\cdot)}  ,
\end{equation}
which is the sum of the torque $\tau_{(\cdot)}^h$, that results from
the force exerted by the user on the handles and transmitted to the
wheel hub through the mechanical structure of the walker, of the term
$- b_{w} \dot \alpha_{(\cdot)}$, accounting for the rolling resistance
that opposes to the wheel rotation (thus having $b_{w}$ as the viscous
friction coefficient around the wheel rotation axle) and of braking
action $\tau_{(\cdot)}^b$.  Notice that pure rolling assumption is
supposed to hold.  Similarly to~\cite{hirata2007motion}, the braking
action is modeled as a dissipative system, i.e.,
\begin{equation}
  \label{eq:Brakes}
  \tau_{(\cdot)}^b = - b^b_{(\cdot)} \dot \alpha_{(\cdot)} ,
\end{equation}
where $b^b_{(\cdot)} \in[0,\,b_{\max}]$ are controllable variables
determining the viscous frictions of the brakes.  Whenever $\dot
\alpha_{(\cdot)} = 0$, to model the fact that the brake has sufficient
authority to keep the wheel at rest, we assume that
\begin{equation}
  \label{eq:Brakes2}
  \tau_{(\cdot)}^b = -c^b_{(\cdot)} \tau^h_{(\cdot)} ,
\end{equation}
where $c^b_{(\cdot)}\in[0,1]$ is an additional controllable variable
acting when $\dot \alpha_{(\cdot)} = 0$. In case of servo brakes,
coefficients $b^b_{(\cdot)}$ and $c^b_{(\cdot)}$ can be changed by
varying the input current, thus allowing the control system to
suitably modulate the braking torque~\cite{hirata2007motion}.
  
By direct experimental measurements made on the system at hand, we
have observed that the torques applied by the user to the mechanical
system are negligible with respect to the maximum braking action.
Moreover, the inertia of the system as well as the maximum forward
velocity $v$ are limited.  As a consequence, the time needed to stop
the wheel rotation is negligible and, hence, it can be assumed
$\omega_{(\cdot)} = 0$ whenever the braking system is fully active.

Due to the described platform, the control law we are designing deals
with a limited set of control inputs.  More precisely, the admitted
control values for each wheel are either $c_{(\cdot)}^b = 0$ and
$b^b_{(\cdot)} = 0$ or $c_{(\cdot)}^b = 1$ and $b_{(\cdot)}^b =
b_{\max}$, i.e., no brake or full brake.  If by chance the brake is
fully active on the right wheel, from~\eqref{eq:WheelDynFinal3} it
follows that $\omega_r(t) \rightarrow 0$.  Hence, the vehicle will
end-up in following a circular path with fixed curvature radius $R =
d/2$, where $d$ is the rear wheel inter-axle length, travelled in
clockwise direction if $v > 0$ (counter-clockwise for $v < 0$).  The
circular path with the same radius $R$ will be instead followed in
counter-clockwise direction if the left brake is fully active and $v >
0$ (clockwise for $v < 0$).  As a consequence, in all the cases of
active braking system, $\omega = v/R$.  If no braking action is
applied at all, the user will drive the {\em FriWalk} uncontrolled.
On the other hand, if both brakes are fully active, the vehicle will
reach the full stop.  From a control perspective, the previous model
turns into a nonholonomic nonlinear vehicle with limited curvature and
quantised inputs: turn left, turn right, move freely or stop.

\subsection{Problem Formulation}
\label{subsec:Problem}

We require the walker to converge to our planned path defined in
$\Pi$, which we will assume to be smooth (i.e., with a well defined
tangent on each point) and with a known curvature.  The planned path
is typically composed of straight segments and circular arcs connected
with clothoids.  The problem to solve is formalised as a classic
asymptotic stability problem:
\begin{equation}
  \label{eq:PathFollowing}
\lim_{t \rightarrow +\infty} \dist(t) = 0,\,\,\,\mbox{and}\,\,\,
\lim_{t \rightarrow +\infty} \te(t) = 0 .
\end{equation}
In general, the dynamic path following problem requires the design of
a stabilising control law $F(t)$ and $N(t)$ for the system
in~\eqref{eq:kin}.  Due to~\eqref{eq:f2t}, \eqref{eq:WheelDynFinal3},
\eqref{eq:Brakes} and~\eqref{eq:Brakes2}, such a solution requires a
proper control low for the control inputs $c_{r}^b$, $b^b_{r}$,
$c_{l}^b$ and $b^b_{l}$.

\section{Approaching the Path}
\label{sec:solution}

The first part of the solution proposed in this paper relies on the
hybrid automaton designed in~\cite{balluchi2005path}, which proposes a
minimum length trajectory to reach a desired path for limited
curvature unicycle-like vehicles.  We first summarise the key points
of this algorithm (for the reader convenience) and then we move to the
description of its generalisation for the problem at hand.

\subsection{Hybrid Solution}

The solution proposed by Ballucchi et al.~\cite{balluchi2005path} is
based on a hybrid feedback controller solving~\eqref{eq:PathFollowing}
for unicycle-like vehicles with bounded curvature radius.  In
particular, the solution there proposed faces the problem of driving a
Dubin's car to a generic path assuming a maximum known curvature for
the path.  The authors show that their hybrid controller is stable
with respect to the path-related coordinates $(\ty, \te)$, where $\ty
= \dist/R$ and $R$ is the fixed maximum turning radius of the vehicle.
The controller automaton comprises three different manoeuvres, i.e.,
{\em Go Straight}, {\em Turn Right} and {\em Turn Left}, which are
defined in terms of the angular velocity $\wt$ of~\eqref{eq:kin} as
\[
\left \{
    \begin{array}{ll}
      \wt = 0 & \mbox{if {\em Go Straight}} \\
      \wt = -\frac{v}{R} & \mbox{if {\em Turn Right}} \\
      \wt = \frac{v}{R} & \mbox{if {\em Turn Left}}
    \end{array} \right .
\]
assuming the forward input $v > 0$ is known.  The solution can then be
straightforwardly mapped onto the control variables available for our
specific problem, i.e.,
\begin{equation}
\label{eq:BangBangBrake}
\left \{
    \begin{array}{ll}
      b_r^b = b_l^b = c_r^b = c_l^b = 0 & \mbox{if {\em Go Straight},} \\
      b_r^b = b_{\max} \wedge c_r^b = 1 \wedge b_l^b = c_l^b = 0 & \mbox{if {\em Turn Right},} \\
      b_l^b = b_{\max} \wedge c_l^b = 1 \wedge b_r^b = c_r^b = 0 & \mbox{if
        {\em Turn Left},} \\
      b_r^b = b_l^b = b_{\max} \wedge c_r^b = c_l^b = 1 & \mbox{if
        {\em Stop}.}
    \end{array} \right . 
\end{equation}

\begin{figure}
  \centering
  \includegraphics[width=.6\columnwidth]{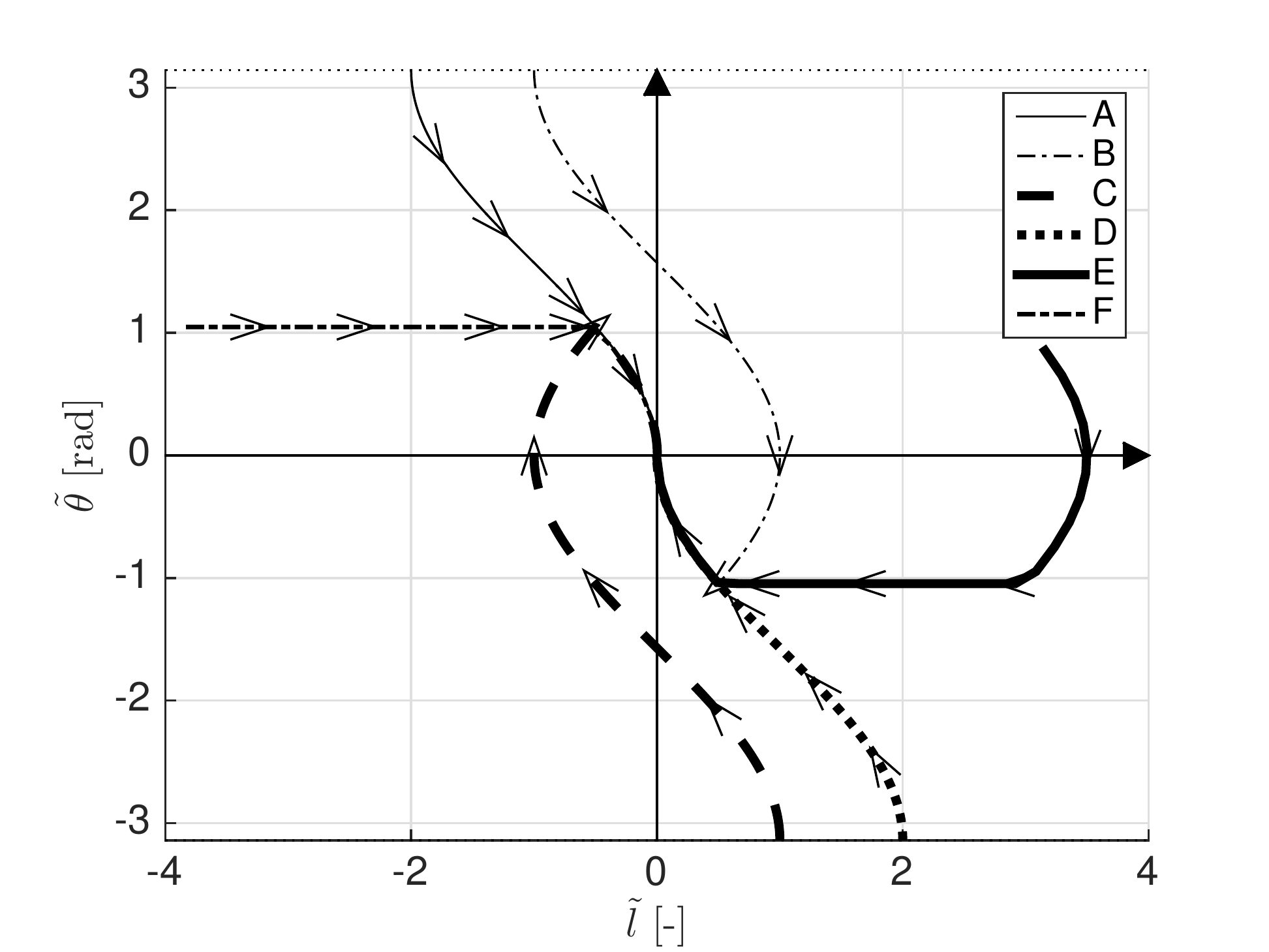}
  \caption{Phase portrait for a set of manoeuvres of the generalised
    hybrid controller~\cite{balluchi2005path} for $\delta = \pi/3$.}
\label{fig:PhasePortBicchi}
\end{figure}

The hybrid automaton comprises three states, in which the three
manoeuvres are coded according to the initial configuration of the
vehicle.  To this end, the state space $(\ty, \te)$ is suitably
partitioned into a set of non-overlapping regions. In each region only
one of the three manoeuvres is active.  For the sake of completeness
we report a typical qualitative trajectory of the robot in the phase
portrait $(\ty, \te)$ in Figure~\ref{fig:PhasePortBicchi}, trajectory
$E$ (solid thick line).
The first part of the manoeuvre is obtained by the {\em Turning
  State}, in which the robot performs a {\em Turn Right} with minimum
radius curvature until it is oriented towards the path to reach (a
linear segment in this example), with $\te = \delta$.  Then the robot
proceeds towards the path in the {\em Straight state} performing a
{\em Go Straight} manoeuvre and, finally, if switches to the {\em
  Controlled state}, in which it rotates performing a final {\em Turn
  Left} manoeuvre and, hence, solving the problem.  Once the robot
reaches the path with the correct orientation, it permanently remains
there providing that the path curvature is feasible according to the
minimum curvature radius constraint.

\subsection{Generalised Hybrid Solution}

\begin{figure}
  \centering
  \includegraphics[width=.6\columnwidth]{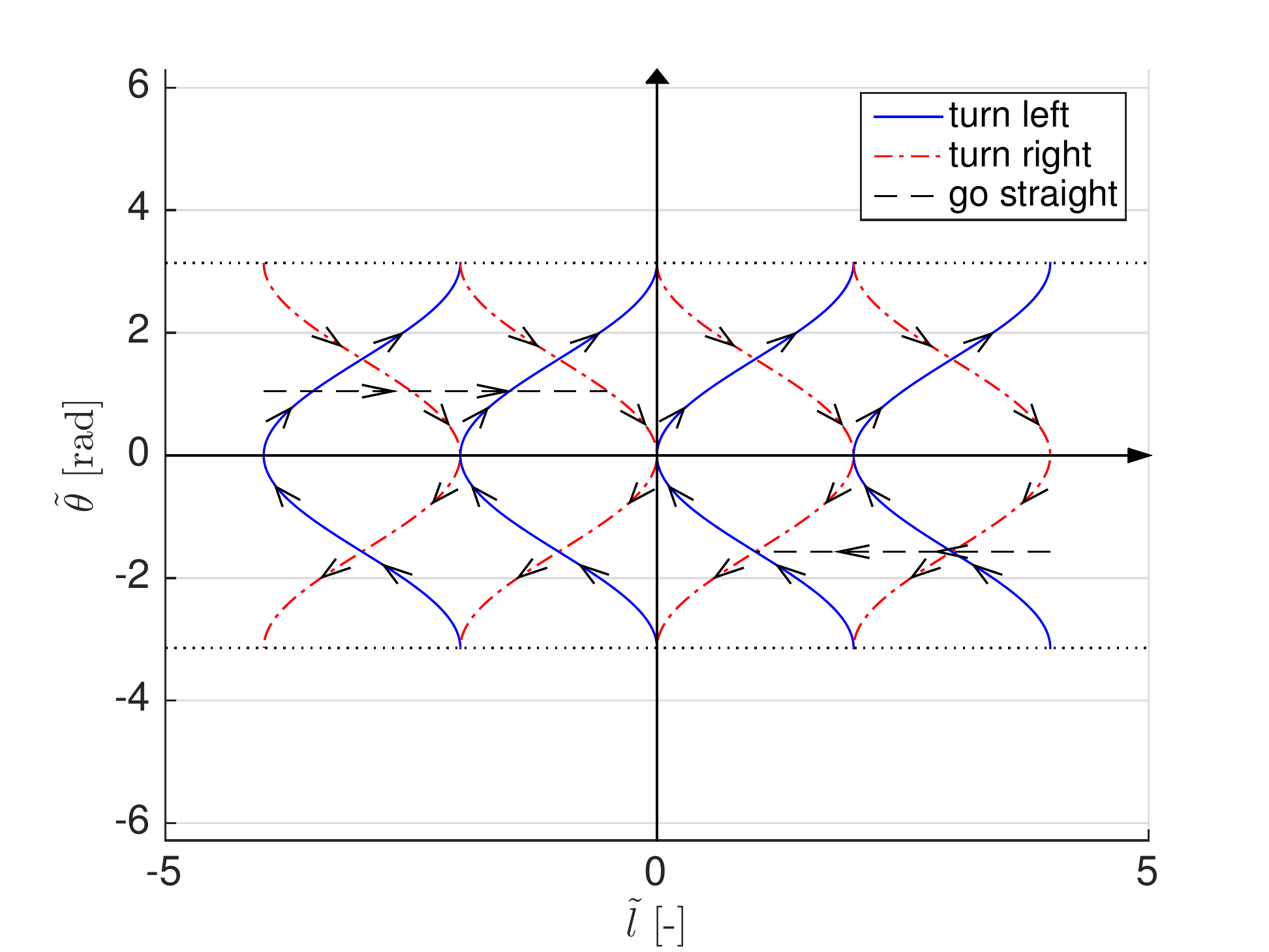}
  \caption{Phase portrait for the {\em Turning State} (solid
    lines). The graph is reported for $\ty = \{-4, -2, 0, 2, 4\}$.
    The dashed line corresponds to the {\em Straight state} for a
    generic choice of $\delta$.}
\label{fig:PhasePortTurning}
\end{figure}

\begin{figure}
  \centering
  \includegraphics[width=.6\columnwidth]{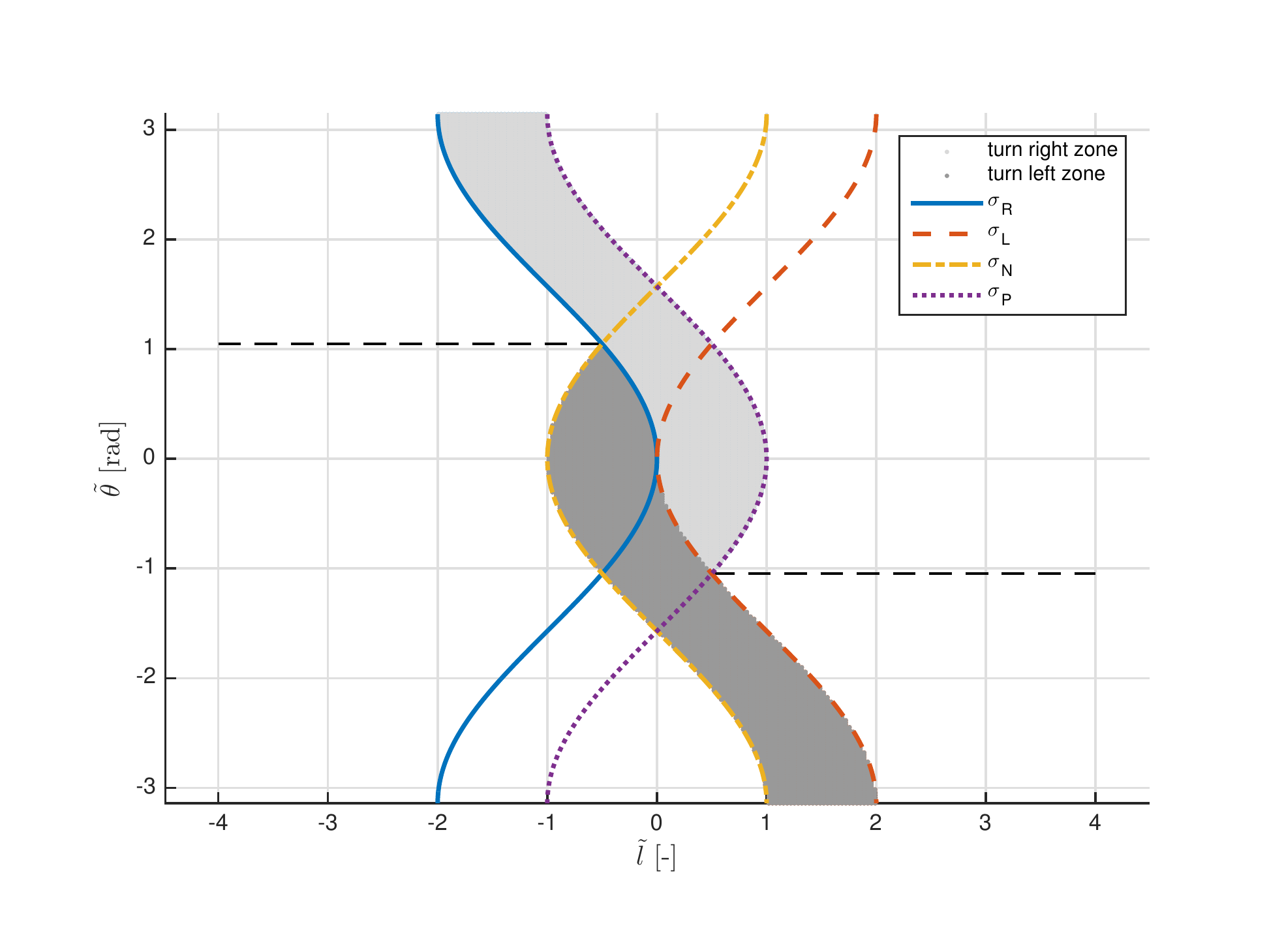}
  \caption{Boundary functions and state space partition induced by the
    boundary function for $\delta = \pi/3$.}
\label{fig:Boundaries}
\end{figure}

In~\cite{balluchi2005path} the choice of the angle of approach $\delta
= \pi/2$ is dictated by the necessity to reach the desired path with
the minimum travelled distance.  Our first generalisation, which is
useful for our specific application since it adds an additional degge
of freedom during the approaching path manoeuvre, is to consider the
angle of approach $\delta$ as a design parameter. In order to extend
the results of~\cite{balluchi2005path} to this broader set of
possibilities, let us first depict in the phase portrait the
trajectories followed by the robot when it continuously rotates on
circles with minimum radius $R$ (Figure~\ref{fig:PhasePortTurning}),
i.e., the trajectories described in the {\em Turning State}.
Notice that circles with any $\bar R > R$, similar graphs are derived.
Furthermore, notice that the angles $\te \in [-\pi, \pi)$.  The dashed
lines in Figure~\ref{fig:PhasePortTurning} represent a possible choice
of $\delta \in (-\pi, \pi)$.  It is now evident that the sequence of
states to reach the path will be given again by: {\em Turning State} -
{\em Straight state} - {\em Controlled state}.  Each of them can be of
zero length.  More precisely, by generalising the boundary functions
reported in~\cite{balluchi2005path} to the case of a generic $\delta$
as:
\begin{equation}
\label{eq:Boundary}
\begin{split}
\sigma_R(\tilde{l},\tilde{\theta}) &= \tilde{l} + 1 - \cos(\tilde{\theta}),\\
\sigma_L(\tilde{l},\tilde{\theta}) &= \tilde{l} - 1 + \cos(\tilde{\theta}),\\
\sigma_N(\tilde{l},\tilde{\theta},\delta) &= \tilde{l} + 1 -2\cos(\delta) + \cos(\tilde{\theta}),\\
\sigma_P(\tilde{l},\tilde{\theta},\delta) &= \tilde{l} - 1 +2\cos(\delta)- \cos(\tilde{\theta}),
\end{split}
\end{equation}
the manoeuvre switching curves are defined, and depicted in
Figure~\ref{fig:Boundaries}.
Those boundaries define the partition of the state space, reported in
the same Figure~\ref{fig:Boundaries} with dark and light grey
colours. In light of this partition, obtained for $\delta = \pi/3$,
from the light grey region the sequence will be {\em Turning State}
(on the right or clockwise) - {\em Controlled state} (turning on the
left or counterclockwise), whose trajectory is named $B$ in
Figure~\ref{fig:PhasePortBicchi} (dash-dotted line).  The same
happens, but with opposite rotations, from the dark grey region ($C$
trajectory in Figure~\ref{fig:PhasePortBicchi}, dashed thick line).
On the left boundary of the light grey region
($\sigma_R(\tilde{l},\tilde{\theta})$ of~\eqref{eq:Boundary}), the
manoeuvre comprises only the {\em Controlled state} (turning on the
right, trajectory $A$ in Figure~\ref{fig:PhasePortBicchi}, solid
line), while from the right boundary of the dark grey region
($\sigma_L(\tilde{l},\tilde{\theta})$ of~\eqref{eq:Boundary}), the
manoeuvre comprises again only the {\em Controlled state} (turning on
the left, trajectory $D$ in Figure~\ref{fig:PhasePortBicchi}, dotted
thick line).  From the dashed line of Figure~\ref{fig:Boundaries}, the
sequence is {\em Straight state} - {\em Controlled state} (trajectory
$F$ in Figure~\ref{fig:PhasePortBicchi}, dash-dotted thick line),
while from all the other position the sequence of states is complete
(trajectory $E$ in Figure~\ref{fig:PhasePortBicchi}, solid thick
line).  It is worthwhile to note that the regions reverted for $v <
0$.  To conclude, we have shown that the generalisation of the
approaching angle to a generic $\delta \in (-\pi, \pi)$ still
preserves the stability property of~\cite{balluchi2005path}.
Nonetheless, it offers a great possibilities for the design of the
control law in the spirit recalled in the Introduction.

\section{Following the Path}

\begin{figure}[h!]
\centering
\subfloat[][\emph{}]
{\includegraphics[width=.45\columnwidth]{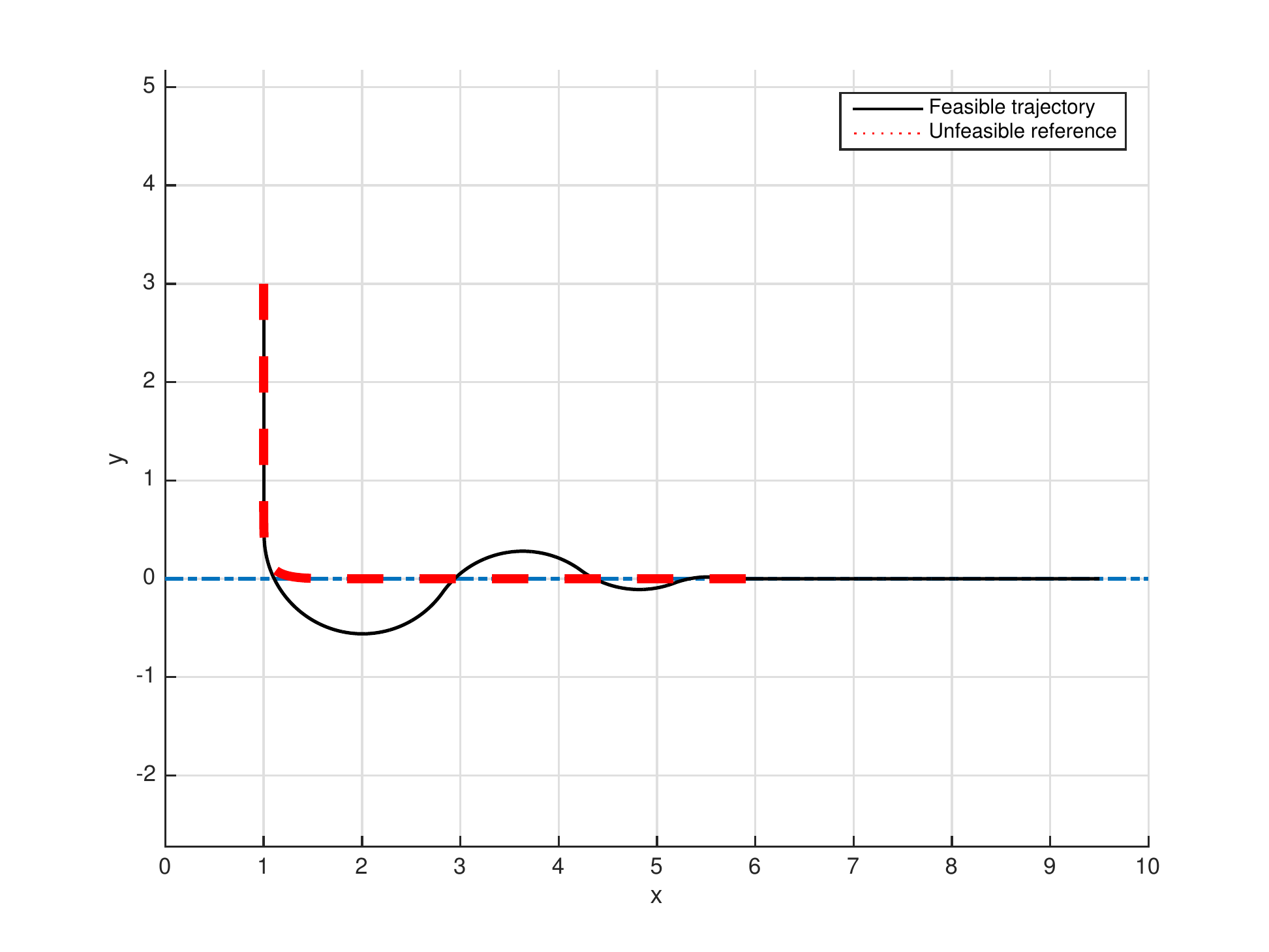}} \qquad
\subfloat[][\emph{}]
{\includegraphics[width=.45\columnwidth]{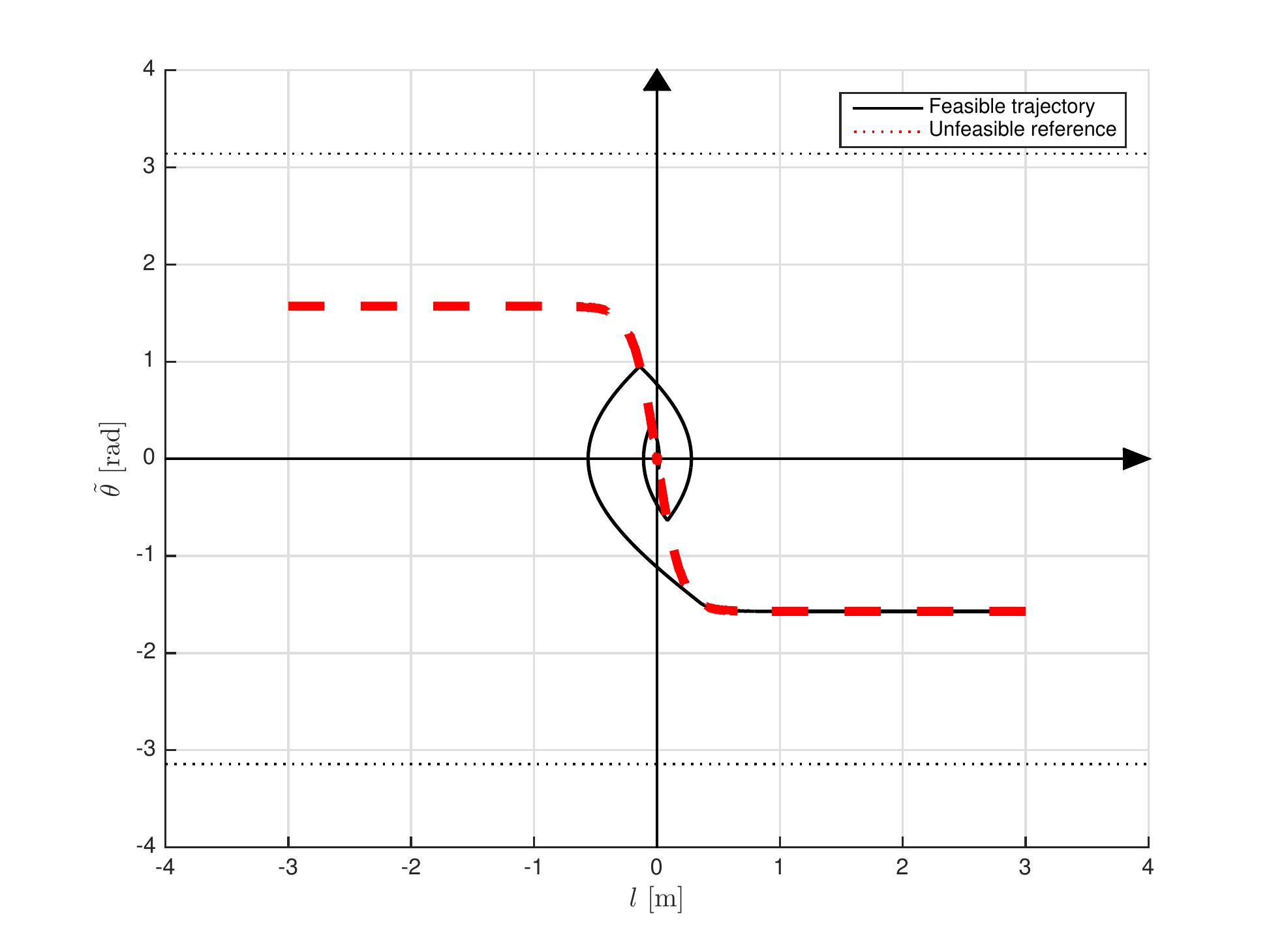}}
\caption{Vehicle trajectory (a) and phase portrait (b) (black lines)
    for an unfeasible reference angle $\delta(\ty)$ (green lines).}
\label{fig:ExampleStab}
\end{figure}

After the approaching phase is completed, the vehicle reaches the proximity of the
path with a minimal set of manoeuvres and with a generic $\delta$.
The solution provided by~\cite{balluchi2005path} to finally stabilise
the robotic platform on the path, even for a generic angle $\delta$,
has two major problems: a) it is perceived as unnatural by the user;
b) if an error on the switching point exists, the vehicle generates a
set of repeated uncomfortable curves, similar to the solid line
reported in Figure~\ref{fig:ExampleStab}.  In order to overcome this limitation and
provide a smooth asymptotic convergence to the path, we present a
further extension of the hybrid automata by considering state
dependent $\delta$ functions.  Having a state dependent approaching
angle to lead the evolution of $\te(t)$ is a common solution for
nonholonomic path following control problems.  Just to mention a few
notable solutions, \cite{SoetantoLP2003} proposes a Lyapunov-based
control law design to govern the torque $N$ in~\eqref{eq:kin} and lead
the vehicle to track the reference angle $\delta(\ty,v)$ defined as an
odd smooth function of $\ty$.  A similar solution based on a nonlinear
system expressed in chained-form has been also presented
in~\cite{jiang2001saturated} for a vehicle with saturated actuation.
The hybrid feedback controller here proposed extends the previous
results on unicycle-like vehicles to quantised control inputs with
customisable approaching angle.  More importantly, it will be shown
that the control law of~\cite{SoetantoLP2003} can be emulated without {\em any}
estimate of the user applied forces and torques and without a precise
modulation of the braking force (contrary to the solution reported in~\cite{FontanelliGPP13})), as reported in
Theorem~\ref{th:StabProof}.  In what follows and without loss of
generality, we focus on $v(t) > 0$, $\forall t$ (therefore,
$\delta(\ty,v)$ simplifies to $\delta(\ty)$).

\begin{theorem}
\label{th:StabProof}
For any function $\delta(\ty)$ continuous, limited
$\delta(\ty)\in(-\pi,\pi)$, with $-\mbox{sign}(\ty) \frac{\partial
  \delta(\ty)}{\partial \ty} \geq 0$ and $\ty \delta(\ty) > 0$,
$\forall \ty \neq 0$, the origin of the space $(\ty,\te)$ is
asymptotically stable.
\end{theorem}
\textbf{\textit{Proof}}
  Let us suppose that the robot is not able to reach the region $\te =
  \delta(\ty)$.  In light of the previous analysis and, in particular,
  by means of the boundary functions~\eqref{eq:Boundary} and the
  region partition of Figure~\ref{fig:Boundaries}, the vehicle is on
  either $\sigma_L(\tilde{l},\tilde{\theta})$ or
  $\sigma_R(\tilde{l},\tilde{\theta})$.  Therefore, the hybrid system
  is in the {\em Controlled state} and the robot reaches $\te =
  \delta(\ty)$ only in the origin.

  Hence, let us suppose that for a certain finite time instant $\te =
  \delta(\ty)$.  If $|\dot\delta(\ty)| \leq v/R$, $\forall \ty$, i.e.,
  the commanded $\delta(\ty)$ has instantaneous curvatures that are
  less than $R$, the vehicle remains on the region $\te =
  \delta(\ty)$ by switching continuously between the {\em Turning
    State} and the {\em Go Straight}.  Notice that the hybrid state
  {\em Controlled state} is never reached, in this case.

  However, it may happen that $|\dot\delta(\ty)| > v/R$ for certain
  values $\ty > \widehat\ty$.  In such a case, the reference of
  $\delta(\ty)$ is unfeasible for the limited turning radius vehicle.
  Let us define with $\bar t$ the time in which the vehicle departs
  from the unfeasible region $\te = \delta(\ty)$ due to the limited
  turning radius $R$. Since $\delta(\ty)$ is a smooth odd function of
  $\ty$ and since the turning manoeuvre is symmetric with respect to
  the $\te = 0$ axis (see Figure~\ref{fig:PhasePortTurning}), when the
  vehicle re-enters into the region $\te = \delta(\ty)$ after a
  turning at time $\widehat t > \bar t$, both $|\ty(\widehat t)| <
  |\ty(\bar t)|$ and $|\te(\widehat t)| < |\te(\bar t)|$ hold.  As a
  consequence, there exist a time $t$ such that $\ty(t) <
  \widehat\ty$, hence $|\dot\delta(\ty)| \leq v/R$ and, finally, the
  previous situation holds.

  We are now able to prove the stability of the proposed feedback
  control law using the Lyapunov stability theory.  To this end, let
  us define the following Lyapunov function candidate
  \begin{equation}
    V(\ty,\te) = \frac{1}{2}\left(\ty^2+\te^2\right) ,
    \label{eq: Lyapunov function}
  \end{equation}
  which is positive definite with respect to the subspace of interest.
  Using~\eqref{eq:kin} it is immediate to note that $\dot V(\ty,\te)$
  is negative definite in the {\em Controlled state} and whenever $\te
  = \delta(\ty)$, which proves the convergence towards the origin from
  any initial condition.

\begin{figure}
  \centering
  \includegraphics[width=.6\columnwidth]{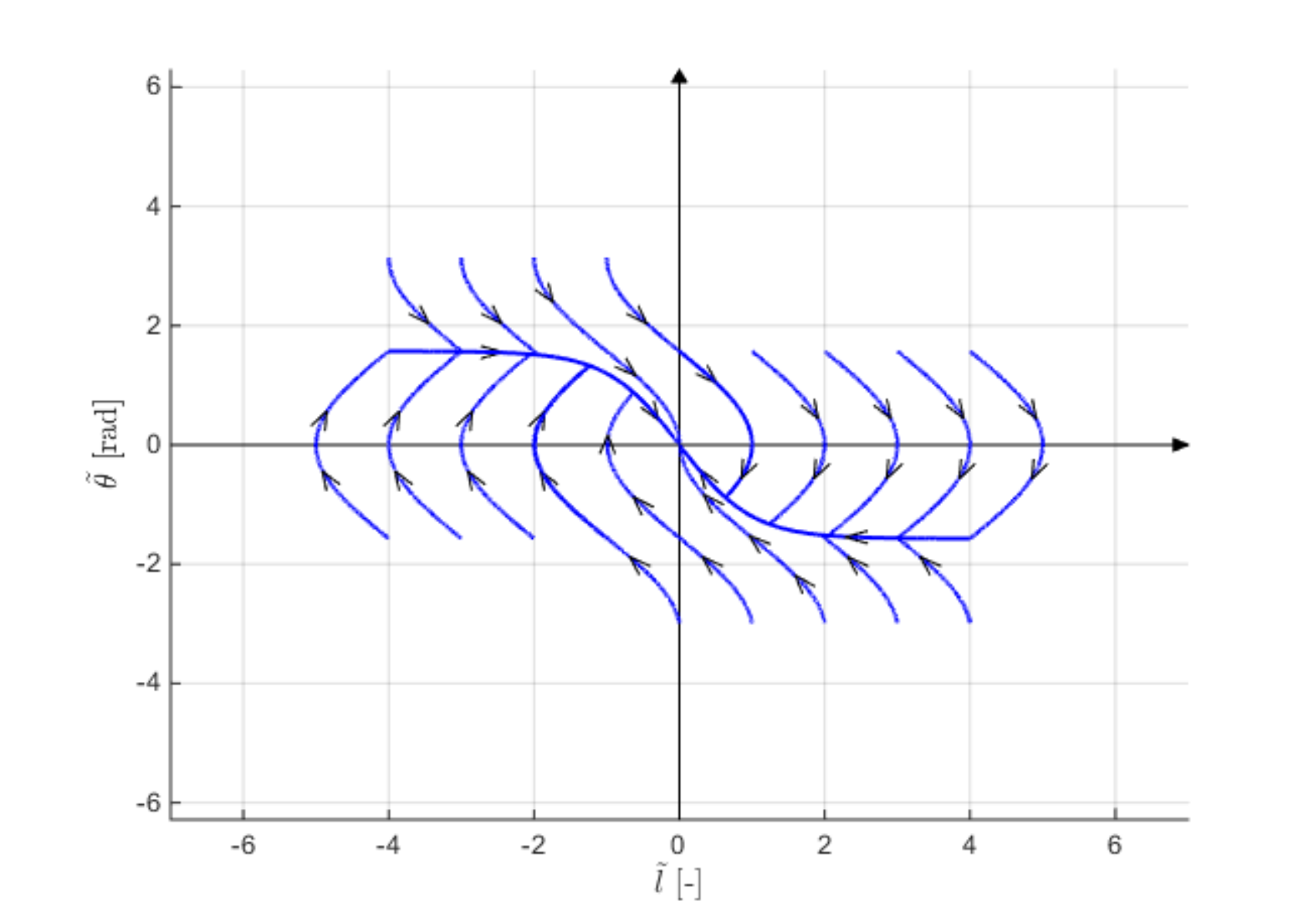}
  \caption{Phase portrait from various initial configurations when
    $\delta(\ty) = \frac{\pi}{2} \tanh(\ty)$ is feasible reference
    angle.}
\label{fig:VarDelta}
\end{figure}

Figure~\ref{fig:VarDelta} depicts the phase portrait from various
initial configurations for $\delta(\ty) = \frac{\pi}{2} \tanh(\ty)$,
i.e., an always feasible approaching angle.
Instead, an example of an unfeasible $\delta(\ty)$ is reported in
Figure~\ref{fig:ExampleStab}, for both the phase portrait and the
Cartesian trajectory followed by the vehicle.

\section{Simulations Results}
\label{sec:results}

\begin{figure}
  \centering
  \includegraphics[width=.6\columnwidth]{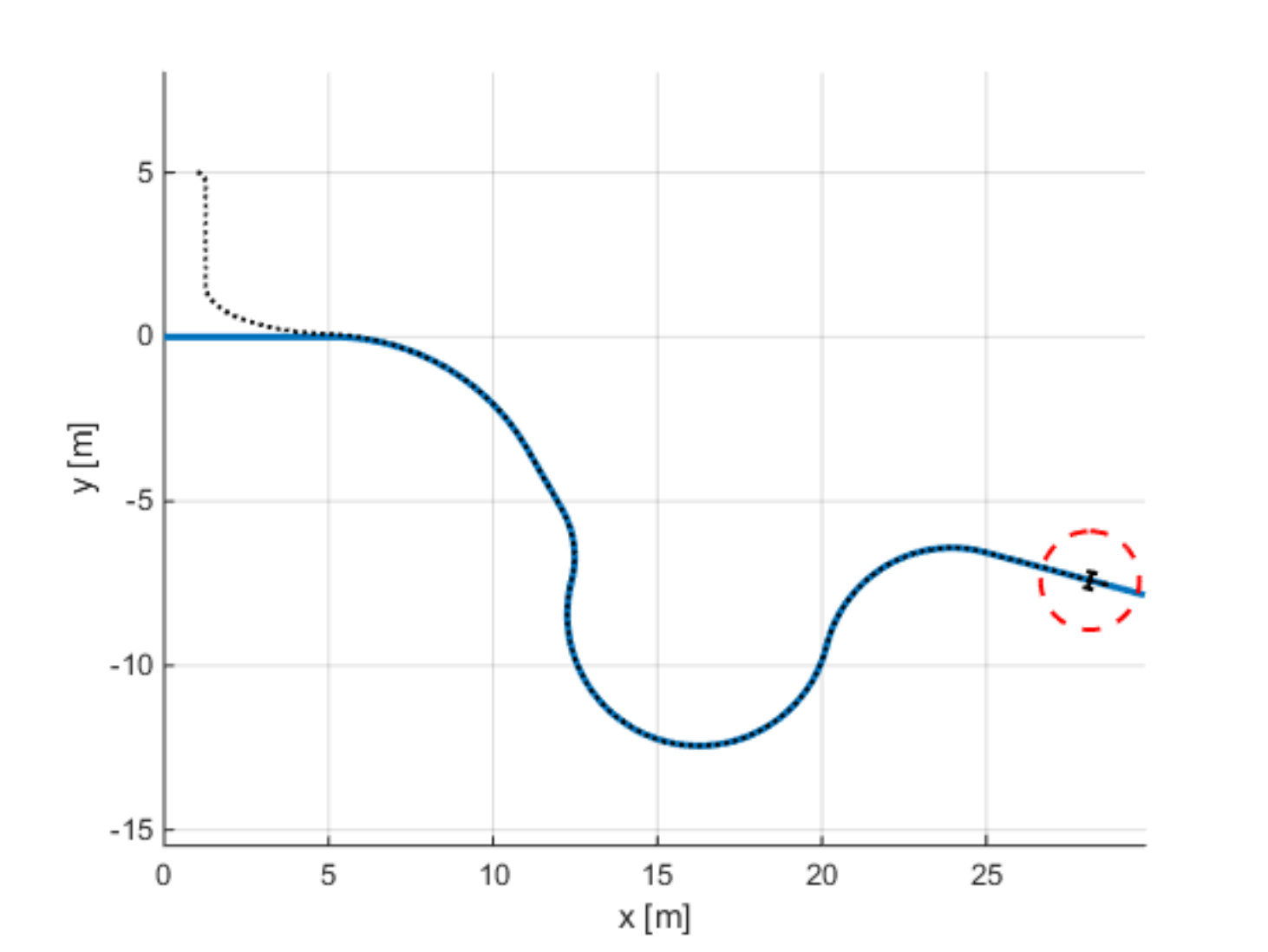}
  \caption{Example of a controlled trajectory.}
\label{fig:PathVehicle}
\end{figure}

\begin{figure}[h!]
\centering
\subfloat[][\emph{}]
{\includegraphics[width=.45\columnwidth]{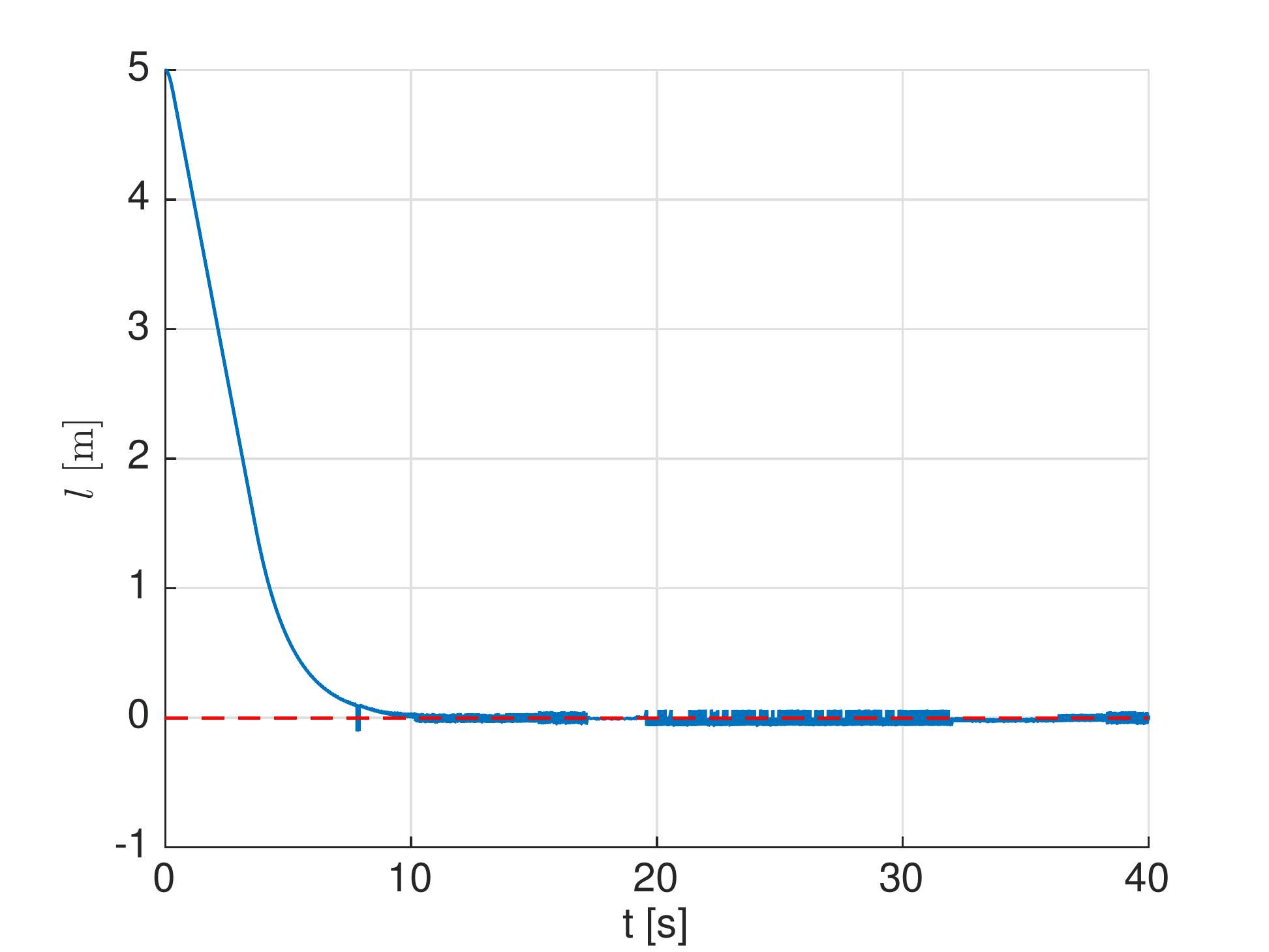}} \qquad
\subfloat[][\emph{}]
{\includegraphics[width=.45\columnwidth]{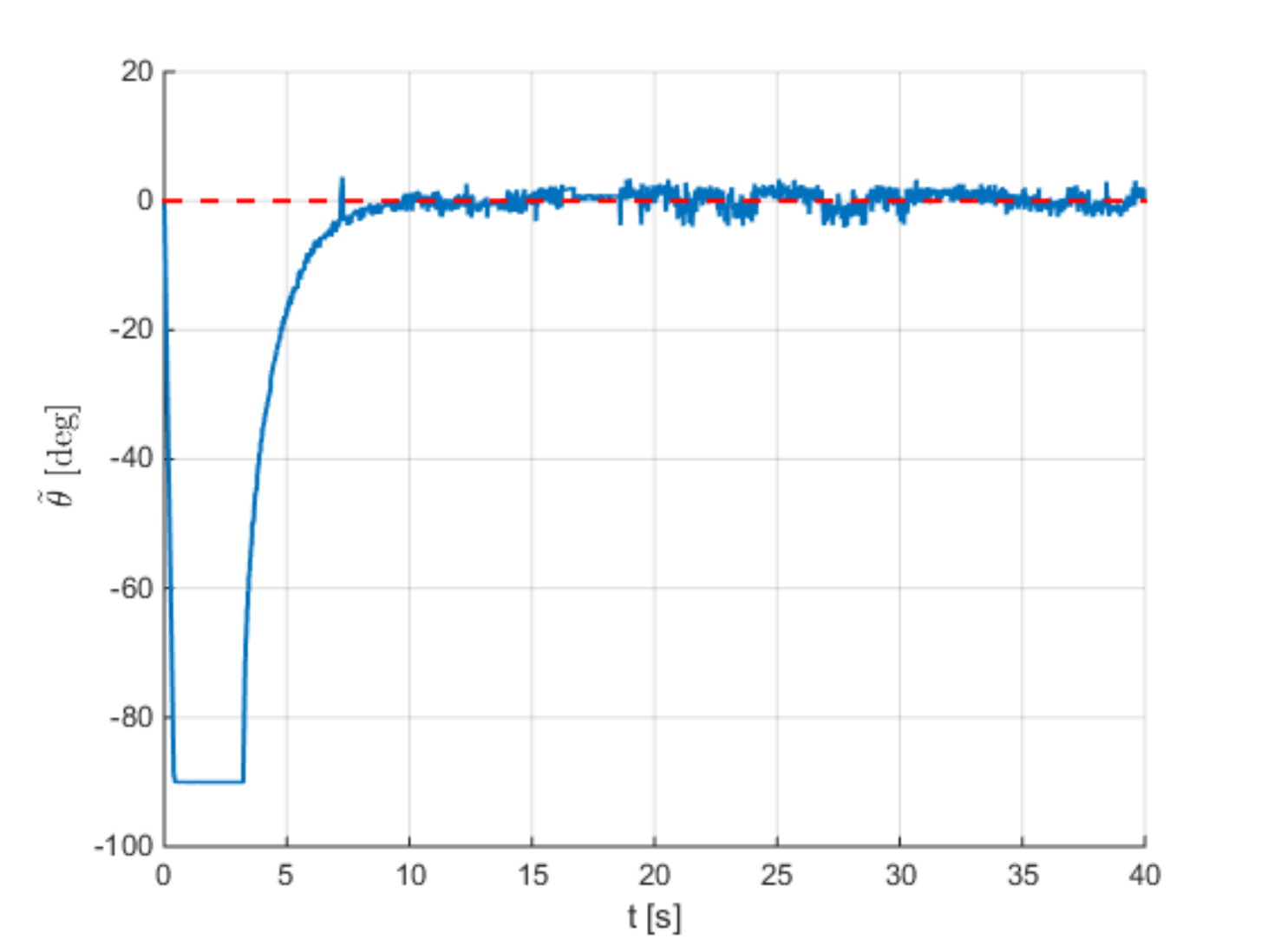}}
\caption{Time evolution of the variables $l$ (a) and $\te$ (b) for the trajectory depicted in Figure~\ref{fig:PathVehicle}.}
\label{fig:Errors}
\end{figure}

To show the effectiveness of the proposed solution, simulations are
reported for a generic path.  Figure~\ref{fig:PathVehicle} reports the
trajectory followed by the vehicle starting with an initial
configuration $(x,y,\theta) = (1,5,0)$.
The final position of the robot on the path is highlighted with a
dashed circle.  The function $\delta(\ty) = \frac{\pi}{2} \tanh(\ty)$ is
adopted.  The distance to the path $l$ and relative angle $\te$ are
reported in Figure~\ref{fig:Errors}.

The high frequency oscillations are due to the continuous switching
between the {\em Turning State} and the {\em Go Straight} hybrid
states, which is necessary to approximate a generic curvature path.

\section{Conclusions}
\label{sec:conclusions}

In this paper we have presented a passive control strategy for a
robotic walking assistant that guides a senior user with mobility
problems along a planned path.  The control strategy exploits the
electromechanical brakes mounted on the back wheels of the walker.
Due to cost limits, the solutions proposed is based on a simple
actuation strategy in which the braking system is controlled with a
bang-bang control.  We show that it is possible to secure a gentle and
smooth path following and to control the vehicle in feedback towards
the desired path by applying two different strategies: when the
platform is far away from the path, we execute an approach phase in
which the walker converges toward the platform with a specified angle;
when it comes in proximity of the platform, the control strategy
switches to a path tracking mode, which uses the four control actions
to converge toward the path with an angle which is a function of the
state.  We have further shown that it is possible to mimic a dynamic
feedback control law without collecting dynamic measures on the
system, hence lowering down its costs.

Future developments will mainly focus on the application of the
proposed algorithm on the actual system and on the reduction of the
annoying switchings among the different states of the hybrid automata
for generic curves.  Another point that serves some further
investigations is related to the user behaviour when it is commanded
to go straight.  In such a case, the strateg should be modified to
account for unreliable and uncooperative users.

\bibliographystyle{IEEEtran}
\bibliography{reference}

\end{document}